\documentclass[letterpaper, 10 pt, conference]{ieeeconf}  

\IEEEoverridecommandlockouts                              

\usepackage{amssymb}
\usepackage{cite}
\usepackage{sdsmmMath}
\usepackage{mdpMath}
\usepackage{roboticsMath}
\usepackage{stemstyling}
\usepackage{graphicx}
\usepackage{subcaption}
\usepackage{pifont}
\usepackage{hyperref}
\usepackage{tabularx}
\usepackage{algorithm}
\usepackage{algpseudocode}
\usepackage{xcolor}

\DeclareMathOperator*{\argmin}{arg\,min}

\newcommand{\failureprob}{\pomdpGenProb{\text{failure}\vert x_{1:N_{points}}, a_{1:T}, \pomdpBelief(\pomdpState_0)}}

\newcommand{\monexl}{\texttt{(M1x,L)}}
\newcommand{\mtwoxl}{\texttt{(M2x,L)}}
\newcommand{\mthreexl}{\texttt{(M3x,L)}}
\newcommand{\POMCP}{\texttt{POMCP}}
\newcommand{\CCPOMCP}{\texttt{CC-POMCP}}

\title{\LARGE \bf
When to Localize?: A POMDP Approach
}

\author{Troi Williams, Kasra Torshizi, and Pratap Tokekar
\thanks{*This research was funded in part by the National Science Foundation (NSF) Eddie Bernice Johnson INCLUDES initiative, Re-Imagining STEM Equity Utilizing Postdoc Pathways (RISE UPP), award \#2217329. All authors are at the University of Maryland, College Park, MD 20742, USA.
        {\tt\small \{troiw,ktorsh,tokekar\}@umd.edu}}
}

\begin{document}

\maketitle
\thispagestyle{empty}
\pagestyle{empty}


\begin{abstract}
Robots often localize to lower navigational errors and facilitate downstream, high-level tasks. However, a robot may want to selectively localize when localization is costly (such as with resource-constrained robots) or inefficient (for example, submersibles that need to surface), especially when navigating in environments with variable numbers of hazards such as obstacles and shipping lanes. In this study, we propose a method that helps a robot determine ``when to localize'' to 1) minimize such actions and 2) not exceed the probability of failure (such as surfacing within high-traffic shipping lanes). We formulate our method as a Constrained Partially Observable Markov Decision Process and use the Cost-Constrained POMCP solver to plan the robot's actions. The solver simulates failure probabilities to decide if a robot moves to its goal or localizes to prevent failure. We performed numerical experiments with multiple baselines.
\end{abstract}

\section{Introduction}\label{sec:introduction}

\begin{figure}[t]
     \centering
     \includegraphics[width=0.75\linewidth]{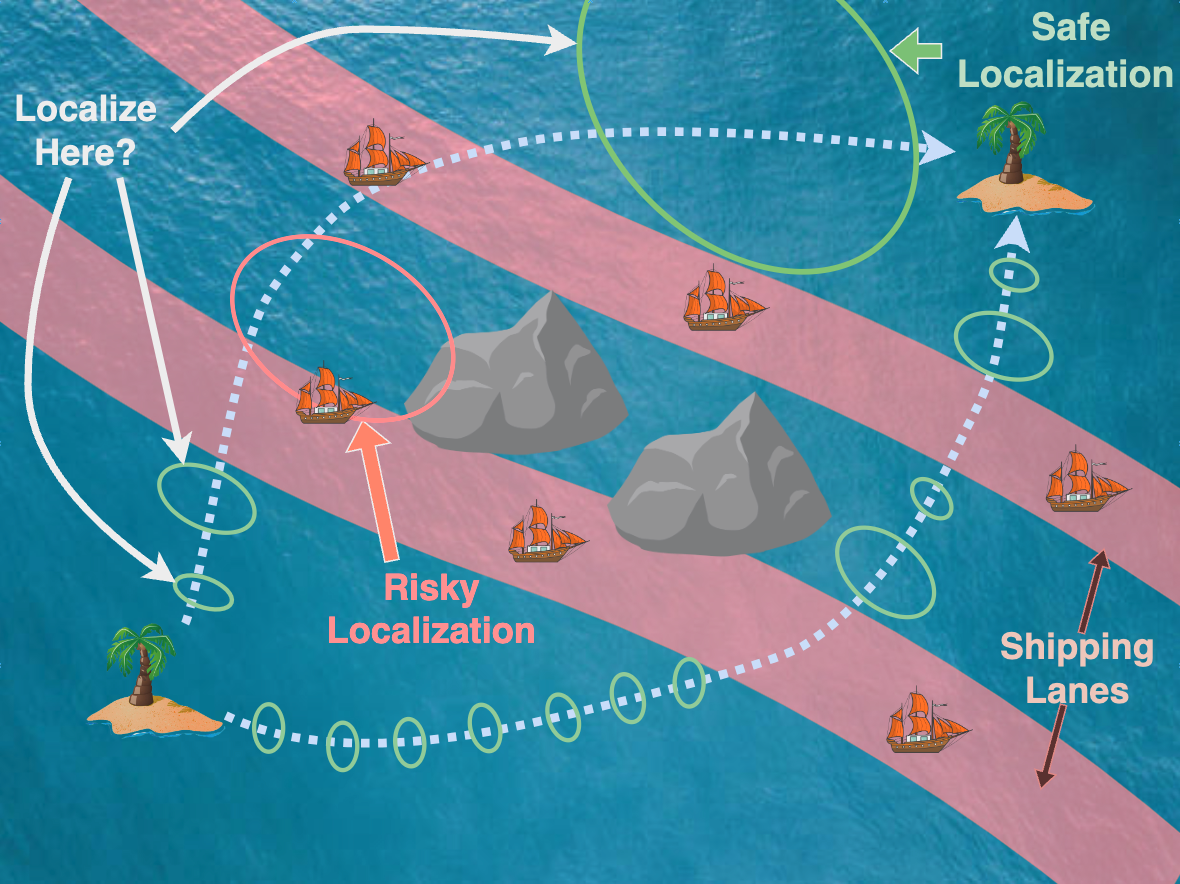}
     \caption{An underwater vehicle follows a path to perform a task (like searching for an aircraft's black box) and must choose when to localize, which requires surfacing and poses a collision risk in shipping lanes. Continuous localization (bottom path) is inefficient yet safe. Selective localization (top path) is more efficient and desirable since the vehicle searches underwater longer. But deciding \textit{when to localize} to avoid hazards and stay on the path is more challenging.}
     \label{fig:w2l_concept}
\end{figure}

Self-localization is crucial in robotics, as it seeds situational awareness and enables a robot to perform downstream tasks such as object manipulation and navigation. In many real-world tasks, an autonomous robot perceives its environment (including self-localization), plans, acts, and then repeats this cycle continuously. However, a robot may seldom want to localize when localization is costly. Let us consider examples where an autonomous underwater vehicle (AUV) navigates through high-traffic shipping lanes \cite{pereira2013risk} to deliver supplies, search for critical items such as black boxes from crashed aircraft, or assess post-tsunami underwater damage (Figure \ref{fig:w2l_concept}). Typically, AUVs localize accurately by surfacing to collect GPS data because underwater dead-reckoning drifts over time, and sonar-based localization is infeasible in featureless environments. However, surfacing risks collisions with ships and increases operating time.

We explore scenarios where continuous localization may be unnecessary or not required, challenging the traditional perceive-plan-act paradigm. Instead, can a robot plan and act for extended durations, localizing only when necessary? Specifically, we aim to address the question: \textit{when should a robot localize} to avoid exceeding the probability of failing a task? We argue that this question is non-trivial. For example, what is the appropriate state uncertainty threshold to trigger localization in a given scenario? Furthermore, should the AUV localize \textit{preemptively} before unsafe regions? If so, how far in advance (\textit{when})?

We propose a move-localize behavior planner based on a constrained Partially Observable Markov Decision-making Process (POMDP) to address such questions. A constrained POMDP allows us to decouple failure probabilities and rewards and reduce ad-hoc reward tuning. Our formulation proposes the probability of failure as a cost constraint with a threshold and creates a reward strategy that penalizes localization. 



\section{Related Work}\label{sec:related_work}

Our work explores the idea ``\textit{when an agent should localize}'', filling a research gap and potentially seeding a novel subarea. To support our assertion, we examine the questions active navigation and path planning tasks have explored and compare them to ours.

Active perception \cite{Cowan1988AutomaticSensor,bajcsy2018revisiting} explores how an agent behaves to achieve one or more goals \cite{Placed2023ActiveSLAMSurvey} and has seeded many active approaches, including active localization, active mapping, and active SLAM. Active localization \cite{Burgard1997ActiveLocalization, Borghi1998MinimumUncertainty} addresses \textit{where an agent moves} and \textit{where it looks} to localize itself \cite{Mostegel2014ActiveMonocular, Otsu2018Where, Gottipati2019DAL, Strader2020PerceptionAware} or a target \cite{Tallamraju2020AirCapRL, Williams2023WhereAmI}. Active mapping, or the next best view problem \cite{Placed2023ActiveSLAMSurvey}, determines \textit{where an agent moves to map} the environment as accurately as possible (for example, \cite{Sasaki2020WhereToMap, Dhami2023PredNBV}). Finally, active SLAM determines \textit{where an agent moves to map and localize} to reduce the uncertainty in its belief and map \cite{Placed2023ActiveSLAMSurvey}. These methods minimize the uncertainty of the agent, its target(s), or the environment, assuming frequent or perfect localization. In contrast, we aim to minimize localization actions while avoiding task failure without guaranteeing minimal agent uncertainty.

Path planning seeks collision-free paths in environments with static or dynamic objects \cite{Mac2016HeuristicApproaches, Karur2021SurveyPathPlanning}, addressing \textit{how an agent moves} between configurations. Path planning methods often assume localization is guaranteed or a robot localizes whenever possible \cite{Tan2010Navigation, Trulls2011AutonomousNavigation, Claes2012Collision, Hennes2012Multirobot, Corominas2020AutonomousNavigation}. Unlike these works, we explicitly plan when to localize to reach our goal.

Tasks related to path planning, such as path selection and dynamic trajectory re-planning, address \textit{which path an agent follows} \cite{Aghamohammadi2014FIRMSF} or \textit{when to deviate} \cite{Agha2014Robust, Aghamohammadi2018SLAPSL,vandenBerg2017MotionPlanning} from a preplanned path. These tasks typically aim to minimize the state uncertainty of the agent, implying frequent localization \cite{Aghamohammadi2014FIRMSF}. In contrast, we aim to mitigate failure, which does not inherently require frequent localization.


\textbf{Contributions.} We introduce an active localization method that decides when agents should localize, contributing a) the first step towards novel methods that decide when an agent should localize, b) a hierarchical planning method based on a constrained POMDP, and c) an open-source implementation \cite{Lee2018MonteCarloTS} of the Cost-Constrained, Partially Observable Monte-Carlo Planning (CC-POMCP) solver, a constrained POMDP solver. 

\section{Problem Statement}\label{sec:formulation}

Our objective is for the robot to safely reach the goal. We seek a policy $\pomdpPolicy=\{a_1,\ldots,a_{N_{actions}}\}$ that 1) minimizes localization events and 2) maintains the probability of failure below a threshold $\hat{c}$ while navigating the path $\boldsymbol{X} = \{x_1,\ldots,x_{N_{points}}\}$. Here, failure includes collisions during localization or entering an unsafe region. This objective is:
\begin{equation}\label{eq:min_localize_policy}
\begin{split}
    \pomdpOptimalPolicy & = \argmin_{\pomdpPolicy\in\Pi} \sum_{t = 0}^{T} \pomdpAction_t = \{\text{localize}\} \\
        \text{s.t.} & ~\failureprob \leq \hat{c},
\end{split}
\end{equation}
where $T$ is the total number of actions, $a_{1:T}$ is the sequence of move and localize actions, and $\pomdpBelief(\pomdpState_0)$ is the initial belief at the start of the path. Finally, $\Pi$ denotes the set of all possible action sequences over $T$ timesteps.

We address objective \eqref{eq:min_localize_policy} using a constrained POMDP. We chose this approach for two reasons: POMDPs have addressed many sequential decision problems (for example, see \cite{Lauri2023Partially}), and constrained POMDPs allow us to separate rewards and costs, simplifying the reward model. We use a constrained POMDP solver to find an action sequence that balances feed-forward (``rarely'' localize) and feedback (localize ``often'') actions, depending on the failure probability.

\section{When to Localize? A POMDP Formulation}\label{sec:method}

We design our planner's behavior using the CC-POMCP solver \cite{Lee2018MonteCarloTS}, employing a hierarchical planning strategy like \cite{Sun2015HighFrequency}. This strategy separates high-level (whether to move or localize) and low-level planning (how to move or localize). Algorithm \ref{algo:active_localize} provides a general description of how our high- and low-level planners control the robot and update the belief as the robot performs actions. The remaining text describes Algorithm \ref{algo:active_localize} in the context of our proposed AUV setup.

\textbf{High-Level Planner (HLP).} The HLP action space is defined as $\pomdpActionSpace = \{\textit{move},~\textit{localize}\}$. If the agent chooses to \textit{move}, we use the transition model to propagate the belief to the next waypoint, a failure state, or a goal state. Then we skip the observation step. If the agent localizes, it surfaces to collect GPS observations, updates its belief, and submerges without moving horizontally. Since surfacing poses a collision risk with boats, the agent will enter a failure state if it localizes within a shipping lane. Finally, regardless of the action, the agent receives a reward and a cost if using CC-POMCP.

\textbf{Lower-Level Planner (LLP).} The LLP determines the sequence of open-loop commands the robot will execute if the action is ``move.'' If ``localize'' is chosen, the LLP makes the AUV surface, update its belief using GPS data, and submerge using no horizontal motion.

\begin{algorithm}[t]
\caption{SafeNav Active Localization Algorithm}
\begin{algorithmic}
\State Initialize High-level planner (HLP) with initial belief $b(s_0)$ and actions $a_t\in\{\text{move}, \text{localize}\}$, and Low-level planner (LLP) with path $x_{1:N_{points}}$.
\State $t = 0$\;
\While {not in a terminal state} 
    \State HLP performs roll-out using $b(s_t)$ to select $a_t$\;
    \If{$a_t = \text{move}$}
        \State LLP computes motion command $u_t$
        \State LLP truncates path, removing current waypoint\;
        \State Robot executes $u_t$\;
        \State HLP uses $u_t$ to compute $b(s_{t+1})$ \;
    \EndIf
    \If {$a_t$ = localize}
        \State Robot receives observation $o_\text{state}$\; 
        \State HLP updates $b(s_t)$ using $o_\text{state}$, producing $b(s_{t+1})$\;
        \State LLP uses $b(s_{t+1})$ to plan hazard-free path to goal\;
    \EndIf
    \State HLP receives reward $r_t$ and, if applicable, cost $c_t$\;
    \State $t = t + 1$\;
\EndWhile
\end{algorithmic}
\label{algo:active_localize}
\end{algorithm}

\textbf{State and Measurement Spaces.} We define these spaces as the agent's position within a grid $\pomdpMeasSpace\subseteq\realset^{2}$. The goal state is located at the end of the robot's pre-defined path. Furthermore, we define the failure states as any location where the agent collides with an obstacle. In the context of an AUV, failure states include collisions with underwater obstacles or trying to surface within shipping lanes. We provide specific details of these failure states in Section \ref{sec:evaluations}.


\textbf{Reward and Cost Models.} The agent receives a reward and cost after executing an action. Our reward model reflects the objective in \eqref{eq:min_localize_policy}: $\pomdpReward_{total} = \pomdpReward_{goal} + \pomdpReward_{move} + \pomdpReward_{local} + \pomdpReward_{fail}$. Here, $\pomdpReward_{goal}$ is the reward for reaching the goal, and $\pomdpReward_{move}$, $\pomdpReward_{local}$, and $\pomdpReward_{fail}$ are the penalties for moving, localizing, or entering the failure state, respectively.

The cost model encodes the constraint in \eqref{eq:min_localize_policy}, setting a cost constraint threshold $\hat{c}$ for the maximum number of particles allowed to enter a failure region. We say ${\hat{c} = p \times N_p}$, where $N_p$ is the number of belief particles and $p$ is the maximum percentage of particles allowed in the failure region.

\section{Numerical Evaluations}\label{sec:evaluations}

The following subsections describe our experimental setup (Section \ref{sec:evaluation:setup}) and results (Section \ref{sec:evaluation:results}). Experiments ran until the AUV reached a goal or failure state, and the results were based on 100 runs for each HLP.

\subsection{Setup}\label{sec:evaluation:setup}
\begin{figure}
    \centering
    \includegraphics[width=0.65\linewidth]{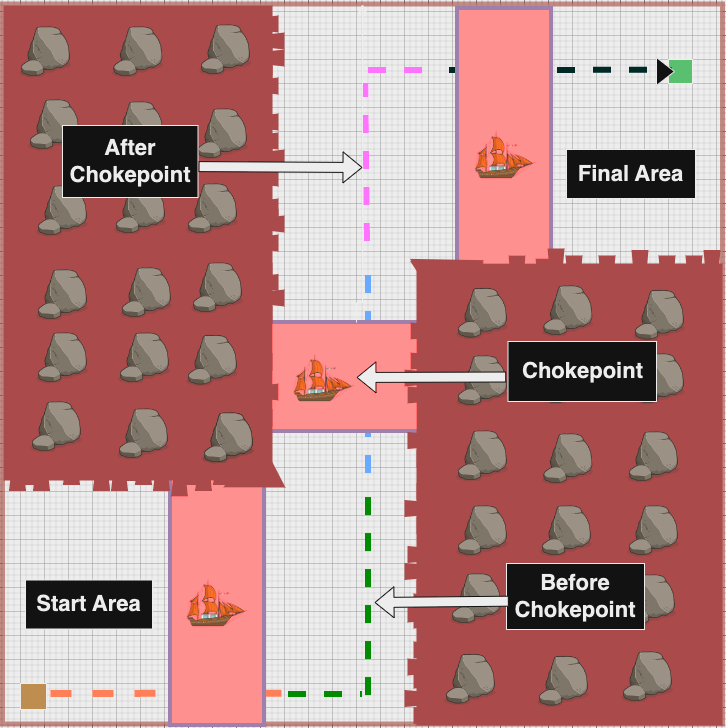}
    \caption{This figure illustrates our training environment (\texttt{ENV-TRAINING}), a $30\times30$ 2D grid world we used to determine the parameters for our online algorithms (shown in Table \ref{table:op_parameters}). The salmon-colored rectangles denote shipping lanes where boats sail while the AUV navigates underwater.
    }
    \label{fig:env_two_large_obs}
\end{figure}

\begin{figure}
    \centering
    \includegraphics[width=0.75\linewidth,trim={0 10cm 0 10cm},clip]{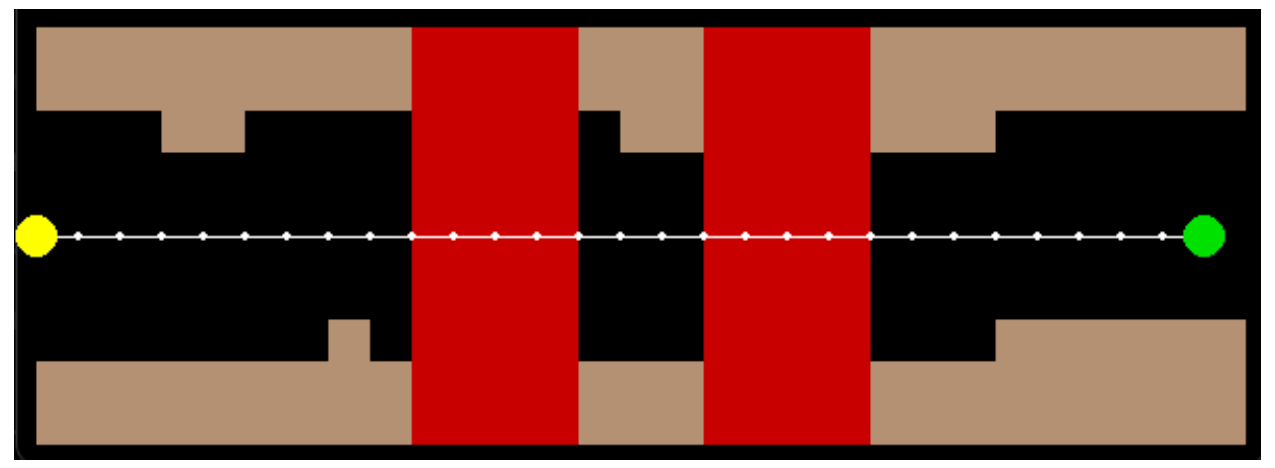}
    \caption{This figure depicts a screenshot of \texttt{ENV-TUNNEL}, one of our evaluation environments. The environment contains obstacles in brown, localize hazards (such as shipping lanes) in red, the (yellow) start and (green) goal positions, and the pre-defined path the AUV should follow.}
    \label{fig:complex_tunnel_envs}
\end{figure}

\begin{figure}
    \centering
    \includegraphics[width=0.75\linewidth,trim={2cm 6cm 16cm 0},clip]{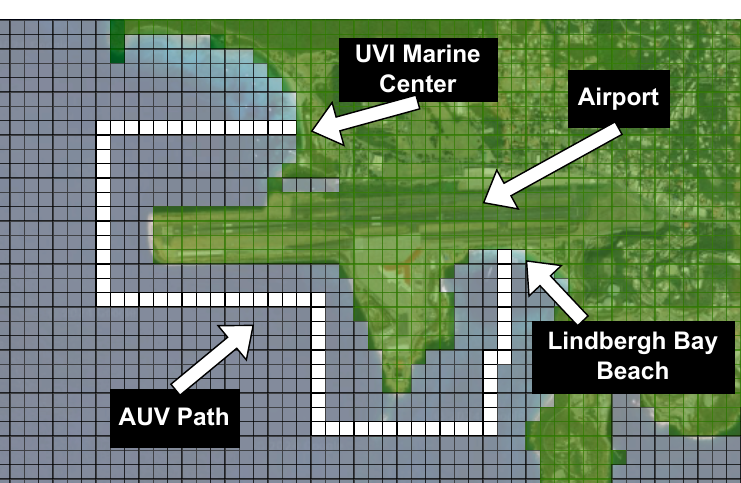}
    \caption{This figure illustrates our real-world evaluation environment called \texttt{ENV-STT}. The image shows a grid overlaying a Google Maps screenshot of the southwestern portion of St. Thomas, U.S. Virgin Islands. The green cells represent obstacles (land masses), the grey cells denote open water, and the white cells represent the AUV's pre-defined path. The AUV starts at the University of the Virgin Islands' (UVI) Marine Center and ends at Lindbergh Bay Beach.
    }
    \label{fig:env_stt_concept}
\end{figure}

\textbf{High-Level Planners (HLPs).} We used three types of HLPs in our experiments. The first type was \textit{offline}, \textit{uniformed}, \textit{cost-unaware} static planners. Here, uninformed means the planners did not use the belief to determine the next action. These static policies (SP) were \monexl{} (that is, move once, then localize), \mtwoxl, and \mthreexl, representing traditional action sequences with varying localization intervals. The second type was \POMCP, an \textit{informed}, \textit{online}, \textit{cost-unaware} planner. We used \POMCP{} to assess the advantage of employing a cost constraint. Finally, the third type was \CCPOMCP, which we used to implement our proposed algorithm. \CCPOMCP{} is an \textit{informed}, \textit{online}, \textit{cost-aware} planner.

\textbf{\POMCP{} and \CCPOMCP{} Details.} We used the \POMCP{} algorithm from the \texttt{pomdp\_py} Python library \cite{zheng2020pomdp_py} and implemented the \CCPOMCP{} algorithm\footnote{Code: https://github.com/troiwill/when-to-localize-pomdp} using the same library. Both algorithms used the same parameters in all experiments, except for the parameters shown in Table \ref{table:op_parameters}. These parameters were derived through a parameter search using the \texttt{ENV-TRAINING} environment. Furthermore, whenever the AUV moved to another grid cell (via a motion command), the AUV had a 94\% chance of moving to the appropriate cell and a 6\% chance of over- or undershooting the cell. We also added noise during particle reinvigoration (the update phase) for both algorithms, which helped mitigate particle deprivation. Finally, we used the GNU Parallel software \cite{Tange2018} to run our online planning experiments in parallel due to each algorithm's high runtime during planning.

\textbf{Low-Level Planner (LLP).} As we mentioned in Section \ref{sec:method}, the LLP determines the low-level motion commands that the robot should perform to go to the next waypoint based on the high-level action. Thus, if the high-level action was move, the LLP determines the next command (that is, up, down, left, or right) using the belief's mean. If the high-level action was localize, the LLP determines if the AUV is still on the path using the belief's mean. If the AUV is not on the path, the LLP finds the nearest waypoint along the path and computes a collision-free path to this waypoint using breath-first search.

\textbf{Environments.} We created one environment (\texttt{ENV-TRAINING}) to find parameters for \POMCP{} and \CCPOMCP{} (Figure \ref{fig:env_two_large_obs}) and two environments (\texttt{ENV-TUNNEL} and \texttt{ENV-STT}) for evaluation (Figures \ref{fig:complex_tunnel_envs} and \ref{fig:env_stt_concept}). The AUV knew the locations of the failure states (obstacles and shipping lanes) and goal states. Finally, we set the initial state of all particles to the start state. 

\begin{table*}
\centering
    \begin{tabularx}{0.905\textwidth} {|c c c c c c c c c c c|}
     \hline
      Planner & $\pomdpReward_{goal}$ & $\pomdpReward_{local}$ & $\pomdpReward_{fail}$& \# Particles & $\alpha_n$ & $\gamma$ & $\kappa$ & Tree Depth & \# Sims & Max \% of Particles Failed\\ [0.15ex] 
     \hline 
     POMCP & 100 & -5 & -10 & 1000 & N/A & 0.999 & 150 & 8 & 2000 & N/A\\
     CC-POMCP & " & -3 & -100 & '' & 0.001 & 0.9 & 200 & " & " & 10\%\\
     \hline
    \end{tabularx}
    \caption{Online Planner Parameters. Here, $\alpha_n$ is the step size, $\gamma$ is the discount factor, and $\kappa$ is the explore constant.}
    \label{table:op_parameters}
\end{table*}

\subsection{Results}\label{sec:evaluation:results}

We compared the HLPs in terms of the physical failure rates, number of localization actions, and cumulative collisions. The SP results represent an ideal scenario, with no path deviations or collisions, serving as a lower-bound performance benchmark for each SP.

\begin{figure*}[t]
    \centering
    \includegraphics[width=0.3\linewidth]{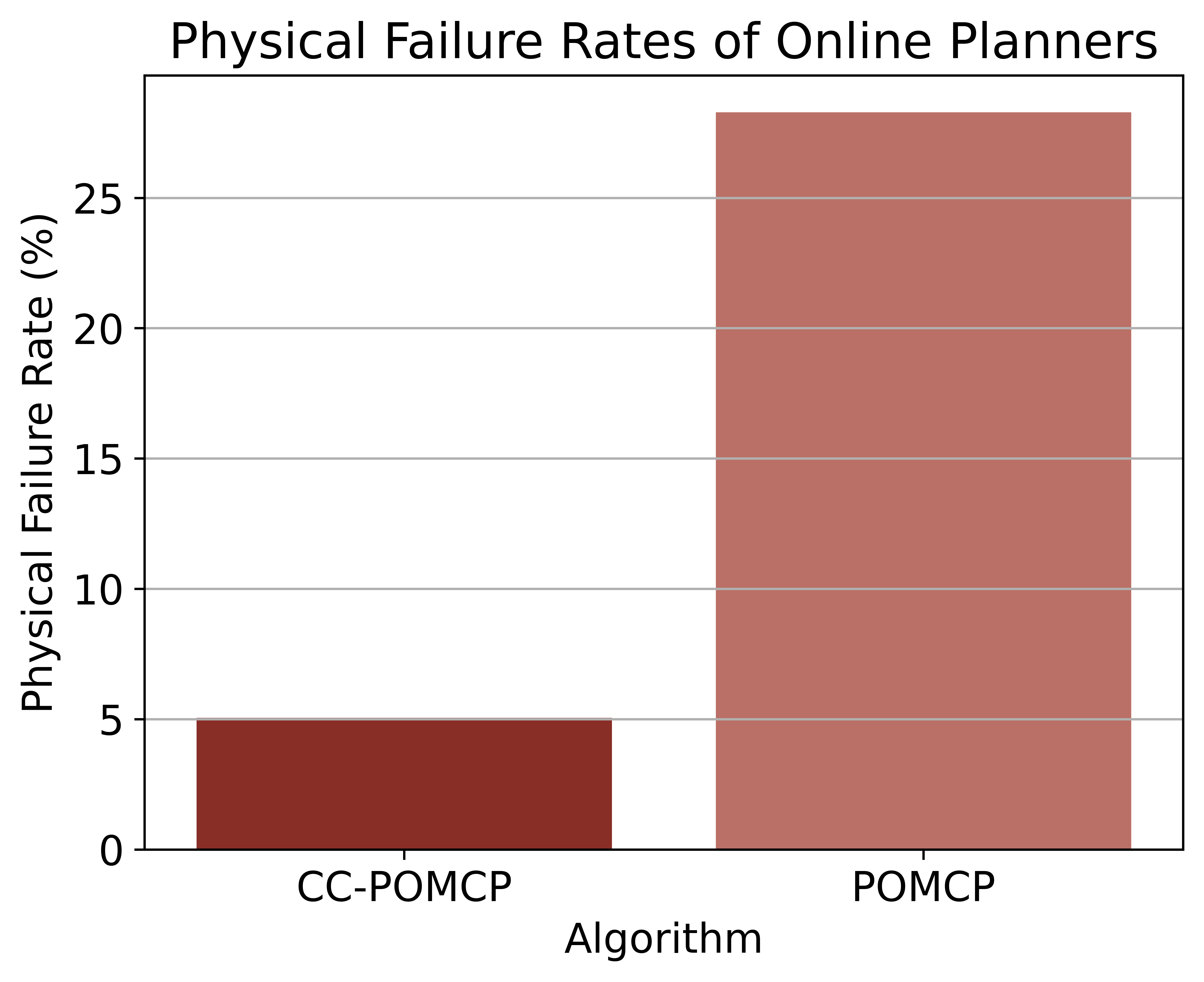}
    \includegraphics[width=0.3\linewidth]{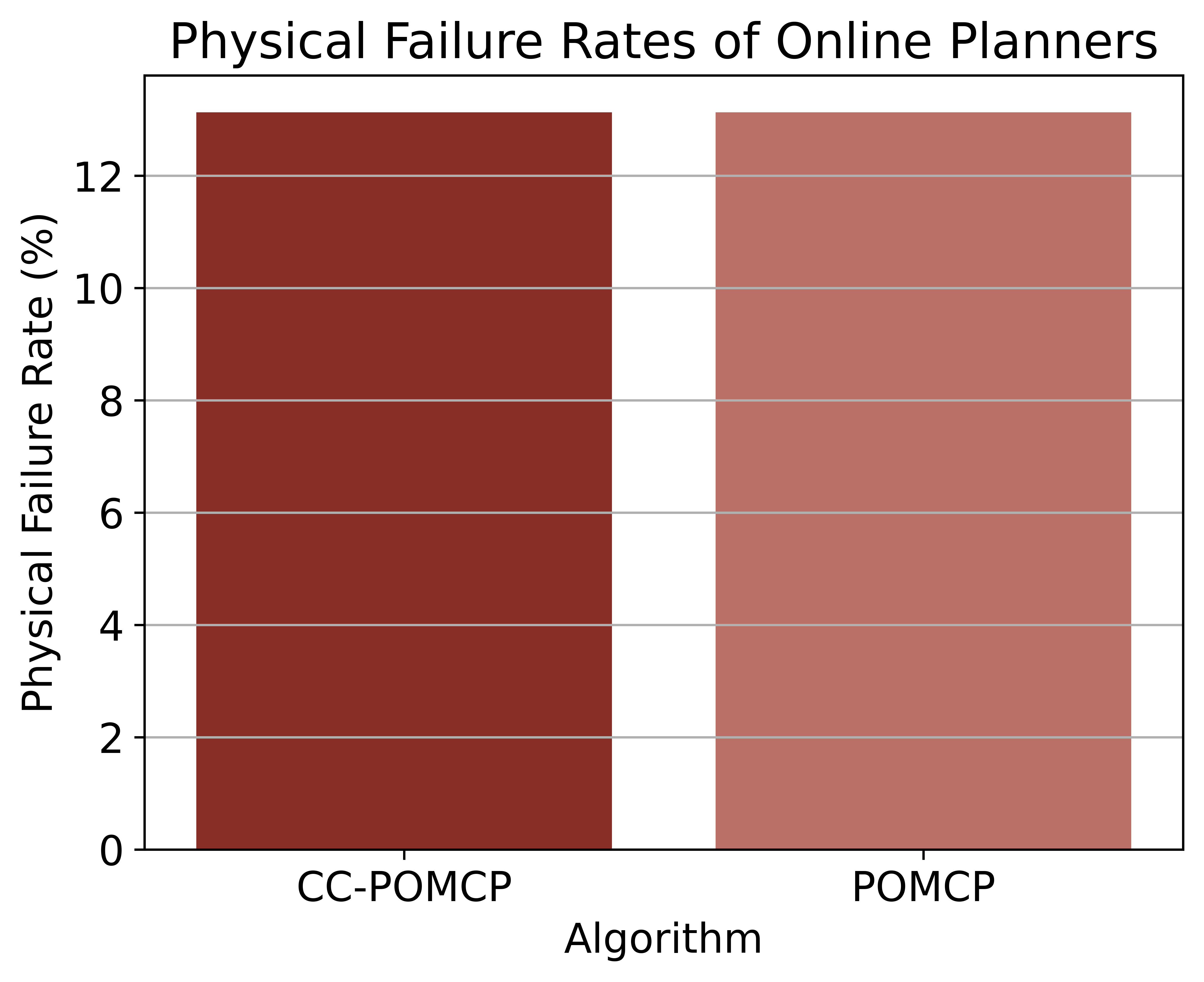}
    \includegraphics[width=0.3\linewidth]{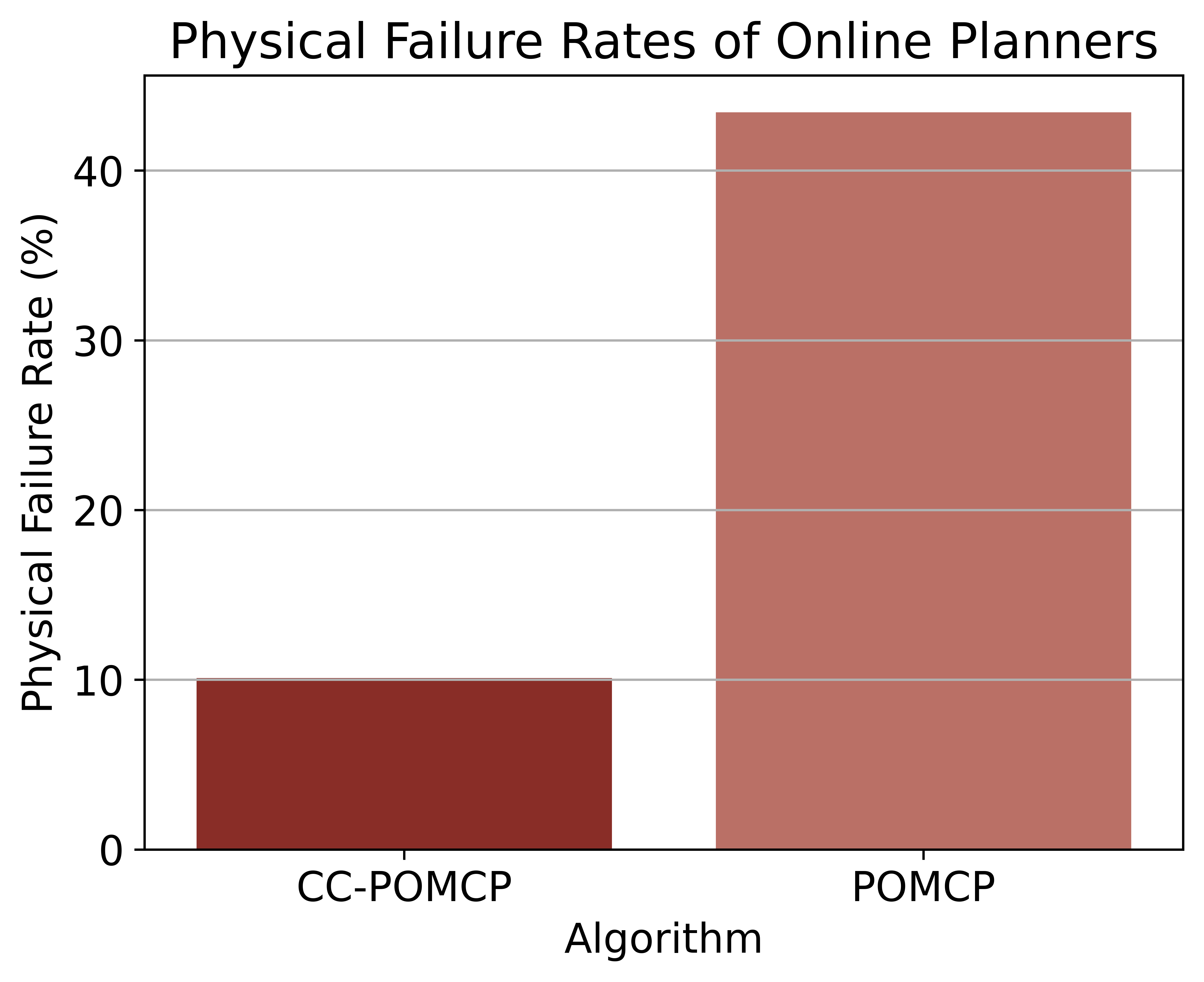}

    \includegraphics[width=0.3\linewidth]{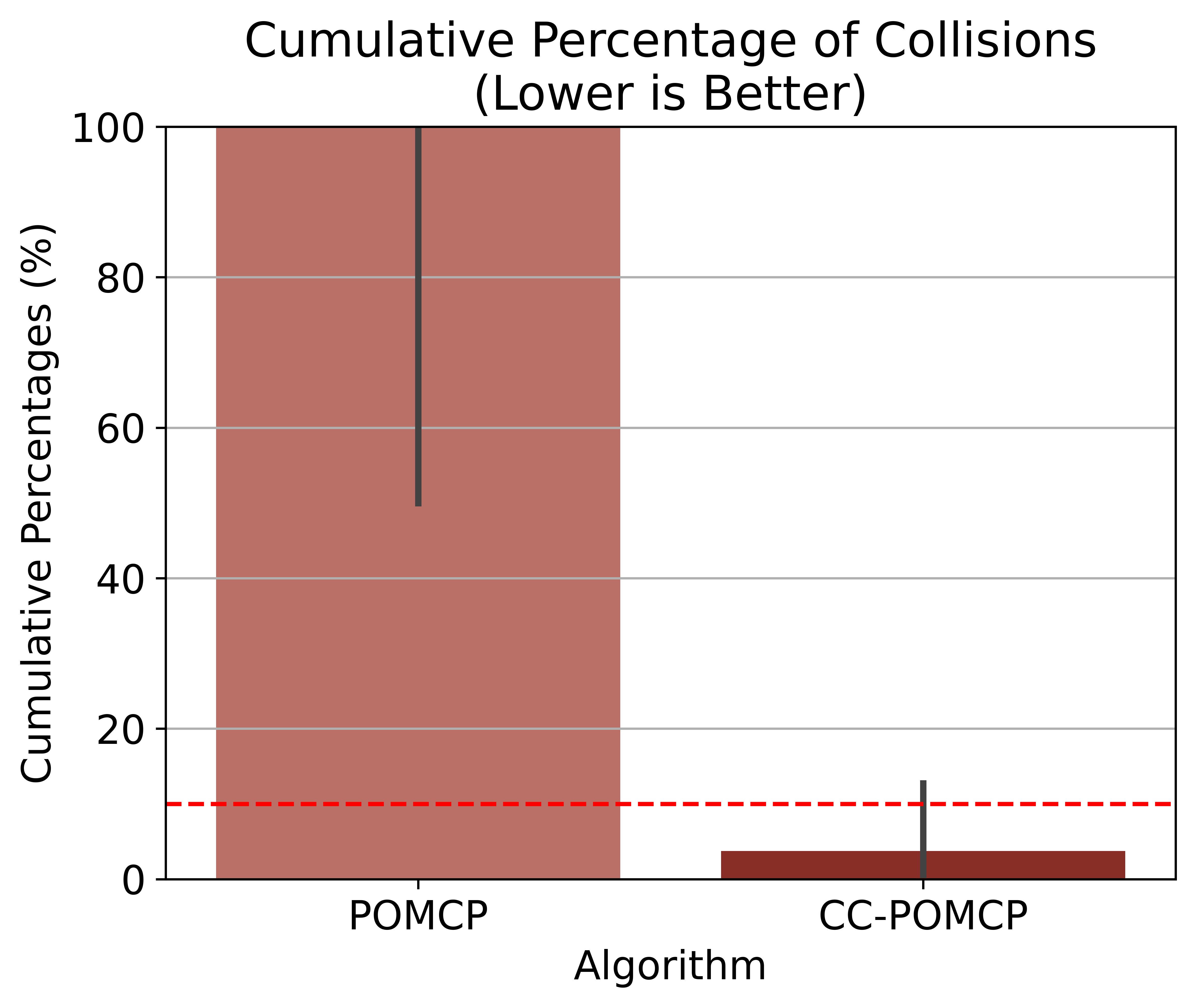}
    \includegraphics[width=0.3\linewidth]{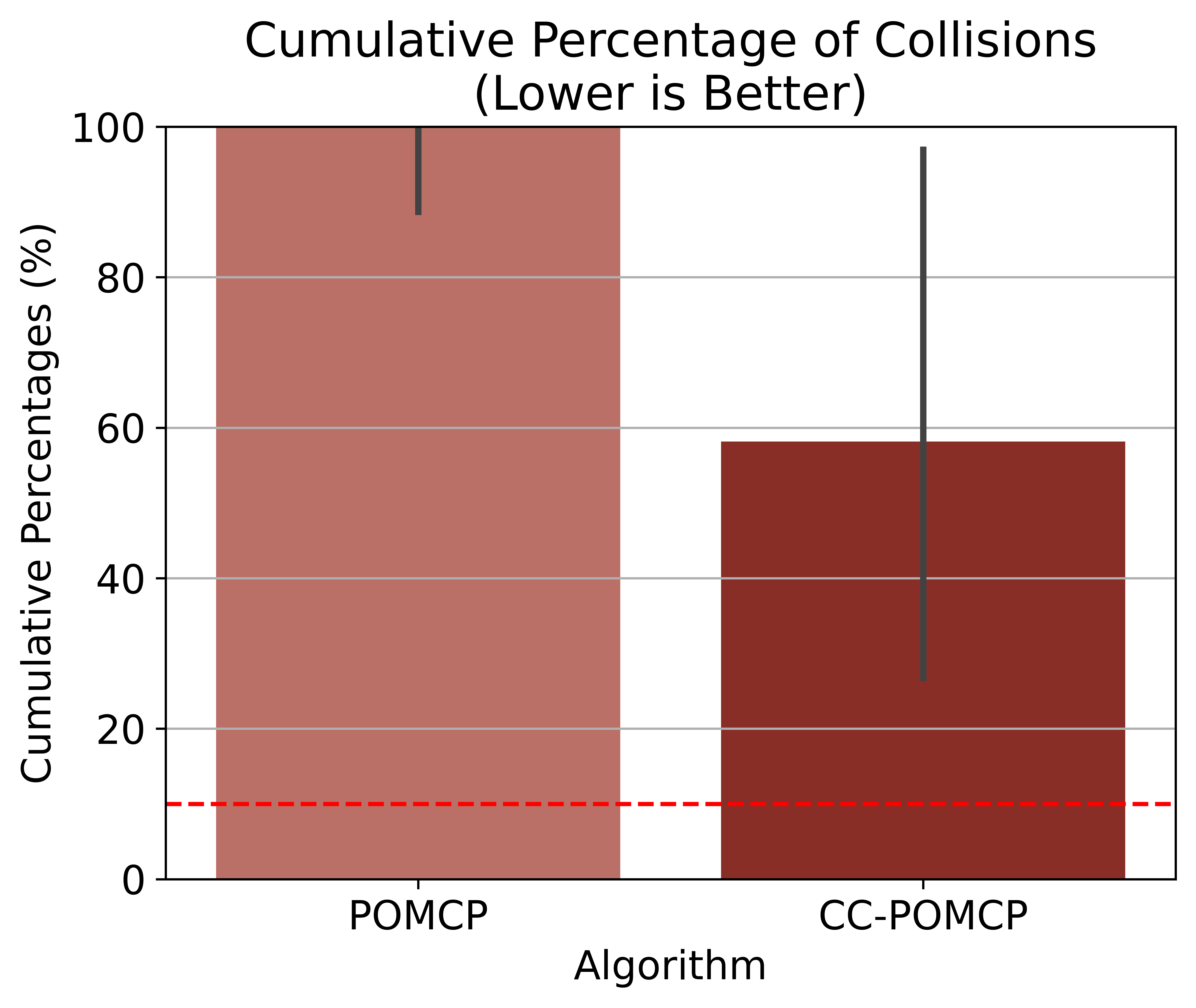}
    \includegraphics[width=0.3\linewidth]{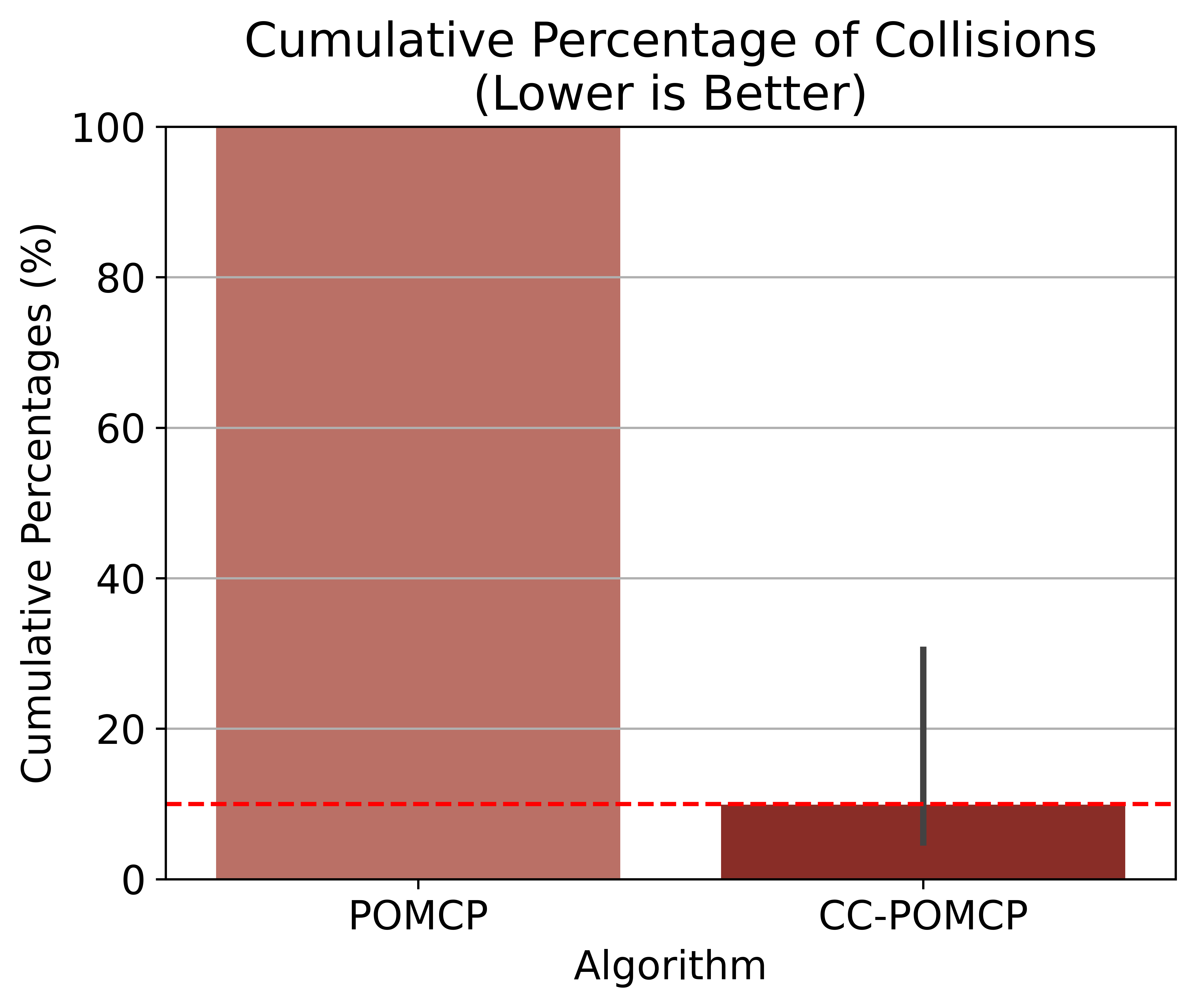}
    \caption{\textbf{The Physical Failure Rates (top) and Cumulative Percentage of Collisions (bottom).} This figure depicts the results for the \texttt{ENV-TRAINING} (left), \texttt{ENV-TUNNEL} (middle), and \texttt{ENV-STT} (right) environments. Here, the failure rates refer to the number of times the robot left the map or collided with an obstacle while moving or localizing. Each graph shows the mean and standard deviation for each algorithm. Due to \POMCP's poor performance, we clipped the cumulative percentages (bottom graphs) to 100\%. Finally, the red dashed horizontal lines in the bottom graphs show the collision threshold of 10\%.}\label{fig:failure_rates_cumulative_perc}
\end{figure*}

\textbf{Physical Failure Rates.} Since the SPs do not consider the environment and belief when planning, they will fail due to collisions within shipping lanes in the \texttt{ENV-TRAINING} and \texttt{ENV-TUNNEL} environments (Figures \ref{fig:env_two_large_obs} and \ref{fig:complex_tunnel_envs}). However, since \POMCP{} and \CCPOMCP{} account for the shipping lanes, no online planner failed due to localizing within a shipping lane. This means that \POMCP{} and \CCPOMCP{} only failed (physically) due to underwater collisions or leaving the map.

Generally, \POMCP{} failed four times or more than \CCPOMCP{} (Figure \ref{fig:failure_rates_cumulative_perc} top). We believe the disparity is due to the difference in the parameters (Table \ref{table:op_parameters}). For example, \POMCP{} has a small failure penalty, which may have caused \POMCP{} to execute more risky policies (such as \textit{never} localize) and ultimately collide with obstacles. Unfortunately, these were the only parameters that enabled \POMCP{} to reach the goal for each environment.

\begin{figure*}
    \centering
    \includegraphics[width=0.4\linewidth]{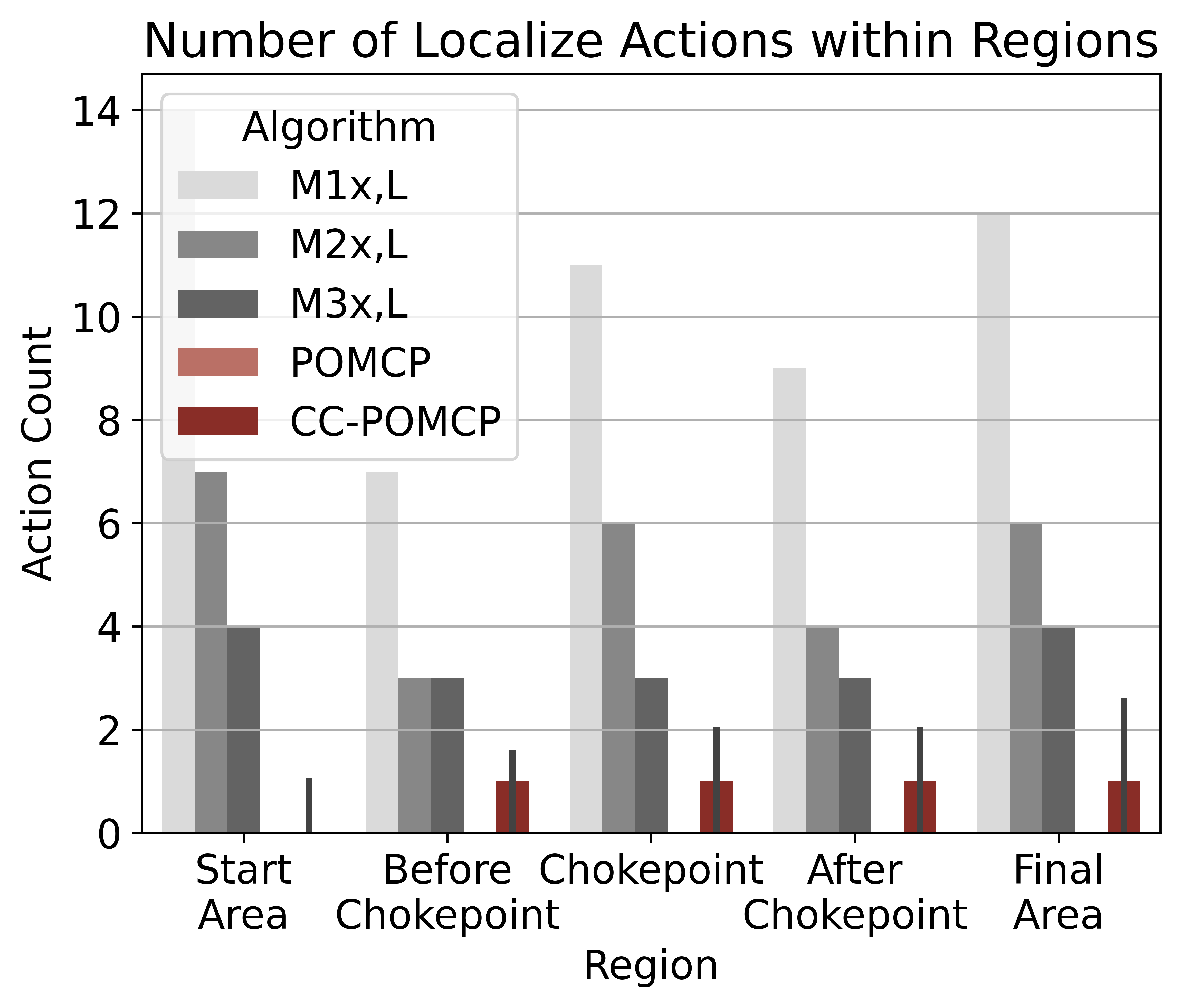}
    \includegraphics[width=0.4\linewidth]{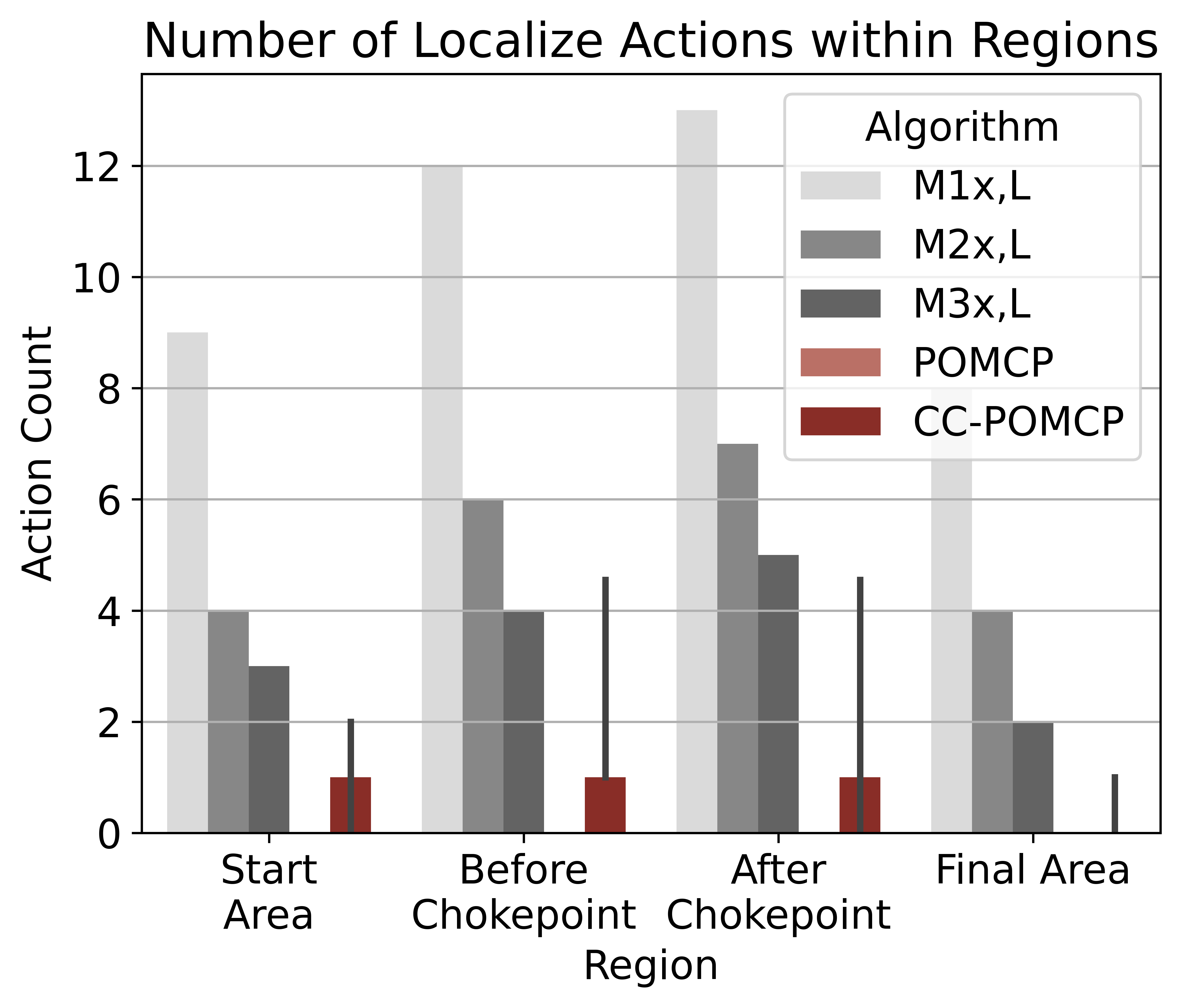}
    
    \includegraphics[width=0.4\linewidth]{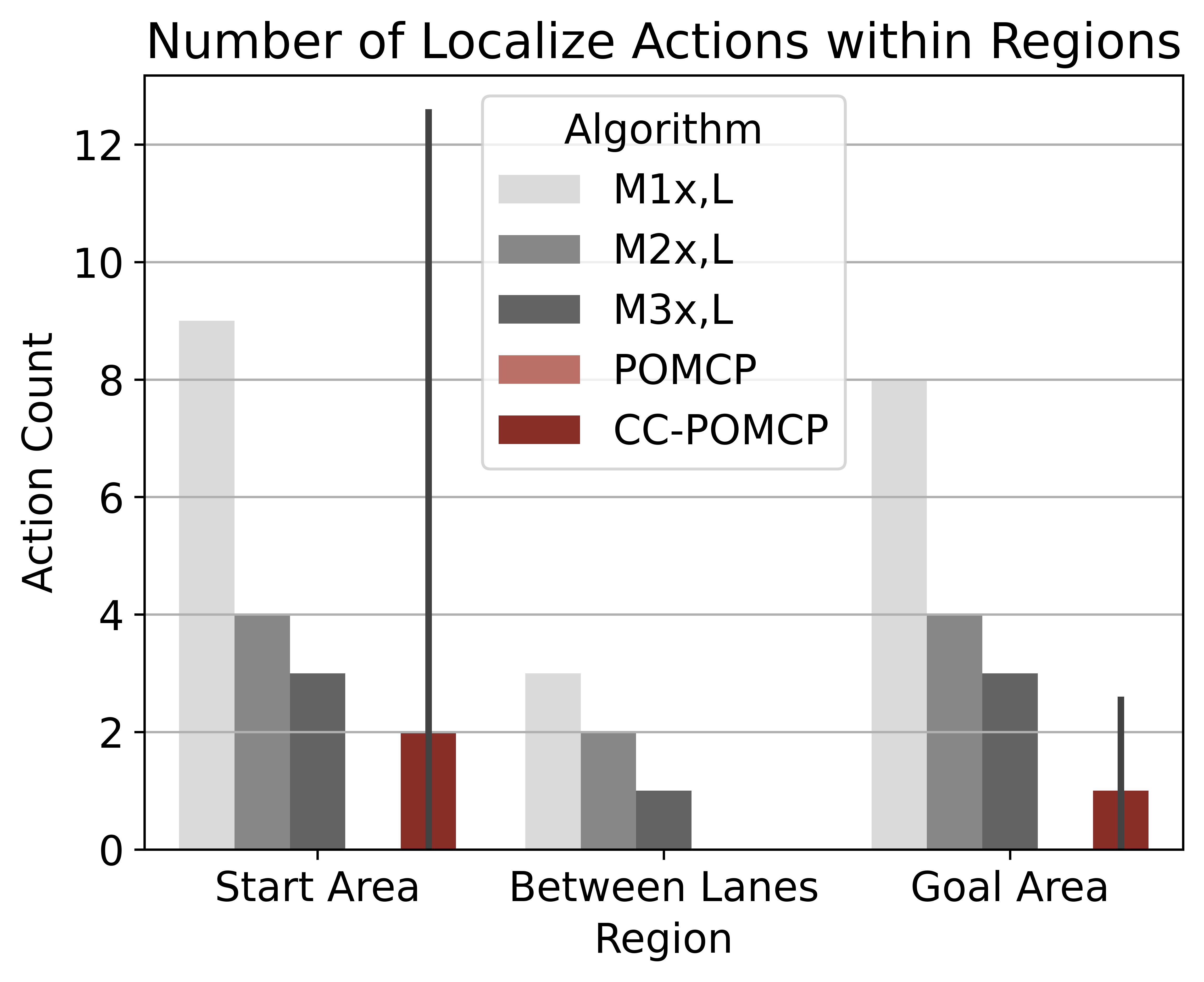}
    \includegraphics[width=0.4\linewidth]{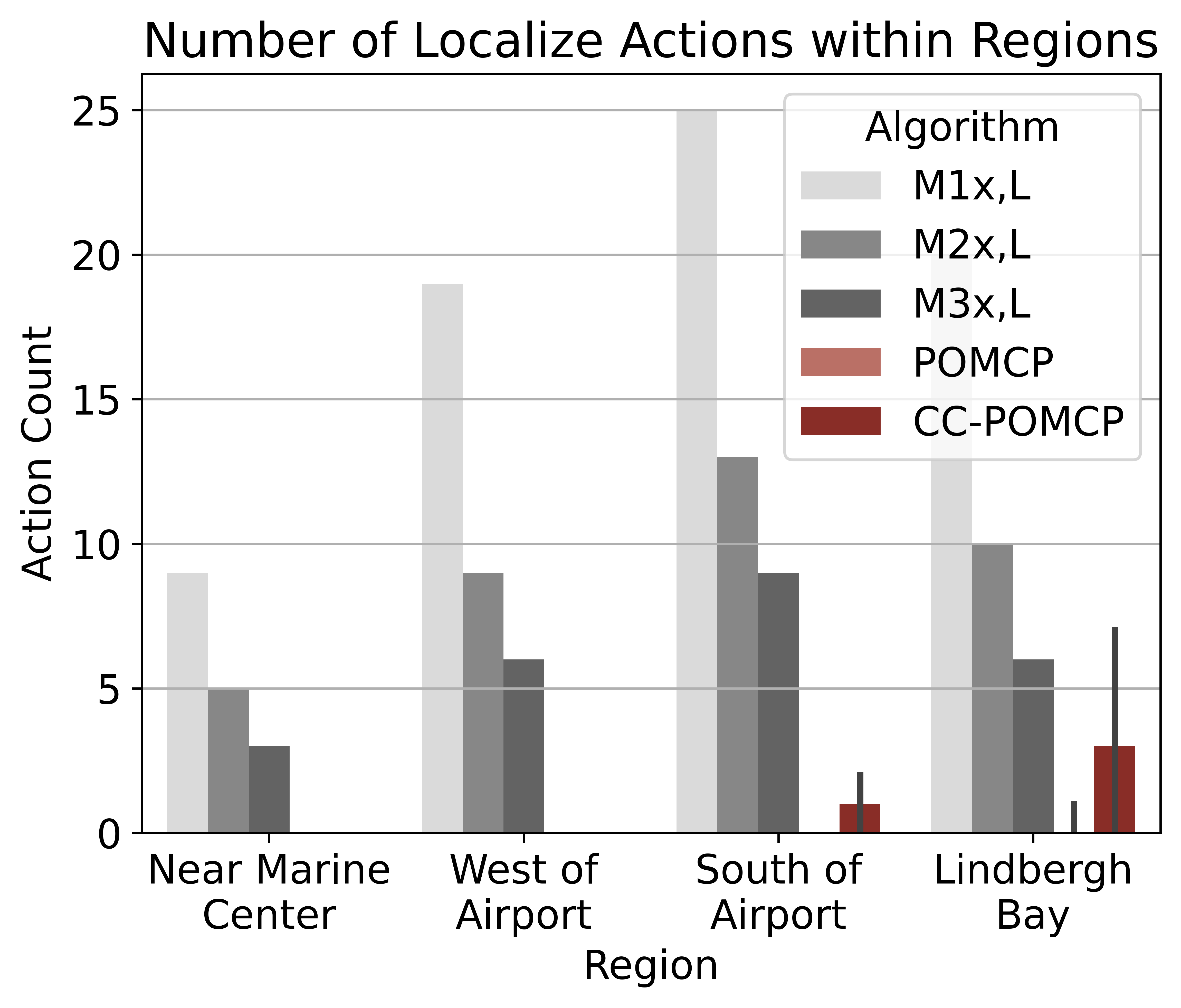}
    \caption{\textbf{Number of Localization Actions for all Algorithms in each Environment's Regions.} The top graphs show the \texttt{ENV-TRAINING} environment results without shipping lanes (left) and with shipping lanes (right). Meanwhile, the bottom graphs show the \texttt{ENV-TUNNEL} (left) and \texttt{ENV-STT} environments (right). The SPs represent the ideal, lower-bound localization counts for SPs. The numbers for \POMCP{} and \CCPOMCP{} represent their empirical values. }\label{fig:localize_count_all_algorithms}
\end{figure*}

In the \texttt{ENV-TUNNEL} environment, both algorithms had similar failure rates (Figure \ref{fig:failure_rates_cumulative_perc}). We believe the rates are similar due to the difficulty of the environment. Figure \ref{fig:localize_count_all_algorithms} (bottom left) shows that \POMCP{} never localized in \texttt{ENV-TUNNEL}, while \CCPOMCP{} localized frequently before the first and after the last shipping lanes. Due to the arrangement of the shipping lanes, both algorithms incurred high uncertainties due to executing 12 or more consecutive motion commands without localizing, increasing the probability of the AUV colliding with underwater obstacles.

\textbf{Localization Actions.} Ideally, the number of localization actions for each SP planner represents the \textit{lowest} number of times a robot would localize in a given environment. Generally, \CCPOMCP{} varied in the number of localization actions executed based on the environment (Figure \ref{fig:localize_count_all_algorithms}). For example, although the average localization counts were similar in \texttt{ENV-TRAINING} (Figure \ref{fig:localize_count_all_algorithms} top), their variances were lower when we performed an ablation study by removing the shipping lanes (Figure \ref{fig:localize_count_all_algorithms} top left). In another example, \CCPOMCP{} avoids localizing between the shipping lanes in \texttt{ENV-TUNNEL} (Figure \ref{fig:localize_count_all_algorithms} bottom left). Instead, \CCPOMCP{} localized prior to the first and after the second shipping lanes. We believe \CCPOMCP{} did this because of the small gap between the shipping lanes and the high belief uncertainty due to multiple open-loop motion commands. Finally, in the \texttt{ENV-STT} environment, \CCPOMCP{} typically localized along the southern part of the airport and relatively more within Lindbergh Bay (Figure \ref{fig:localize_count_all_algorithms} bottom right). We believe \CCPOMCP{} does not localize near the Marine Center or along the western part of the airport because a) there are less number of obstacles near the AUV's path and b) the AUV's uncertainty always starts small in the beginning areas. Such observations show the benefit of online planning versus static planning. That is, we localize more often to avoid collisions in the future and significantly less when collisions are either not an issue or highly probable.

\textbf{Cumulative Collisions.} \CCPOMCP{} outperformed \POMCP{} in terms of the cumulative collisions. In all three environments, \POMCP{} exceeded the collision (cost) threshold by a significant amount (Figure \ref{fig:failure_rates_cumulative_perc}). This result is unsurprising because \POMCP{} is a cost-unaware algorithm. Furthermore, the failure penalty, which influences how often the AUV localizes, may be too small in magnitude. As a result, one must tune the failure penalty in an indirect, ad hoc manner to prevent \POMCP{} from exceeding the collision threshold.

On the other hand, \CCPOMCP{} did not exceed the collision threshold in the \texttt{ENV-TRAINING} (left) and \texttt{ENV-STT} (right) environments on average. However, their variances illustrate that the algorithm did exceed the threshold at least once for both environments. Finally, we note \CCPOMCP's average cumulative percentage does exceed the threshold in the \texttt{ENV-TUNNEL}. Again, we believe the poor performance is due the difficulty of the environment. Here, the AUV cannot localize for an extended period, causing more particles to collide with underwater obstacles. These observations imply that a cost-aware planner is more beneficial than a cost-unaware planner.

\section{Conclusion}\label{sec:conclusion}
We developed a behavior planner for robots to determine \textit{when to localize}, aiming to threshold probabilities of failure (such as collisions). This approach differs from traditional robotics questions (Section \ref{sec:introduction}). Our results showed two key findings. First, online planners allow robots to localize as needed, potentially achieving higher rewards, especially when localization is costly. Second, cost-aware planners provide a precise mechanism for setting performance thresholds like failure probabilities for collisions, reducing arbitrary reward tuning.

One limitation of our work is that CC-POMCP has a high compute time and needs for well-defined transition and observation models. We addressed these limitations by replacing CC-POMCP with a reinforcement learning approach that employs particle filtering and a recurrent Soft Actor-Critic network \cite{shek2024localizeriskconstrainedreinforcementlearning}.

Future work will involve conducting more extensive experiments, including real-world experiments, and relaxing assumptions. For example, we will assume heteroscedastic measurement noise for more realistic scenarios, as in \cite{williams2019learning, williams2021learningA, Williams2023WhereAmI}.

\addtolength{\textheight}{-6cm}   





\bibliographystyle{IEEEtran}

\begin{thebibliography}{10}
\providecommand{\url}[1]{#1}
\csname url@samestyle\endcsname
\providecommand{\newblock}{\relax}
\providecommand{\bibinfo}[2]{#2}
\providecommand{\BIBentrySTDinterwordspacing}{\spaceskip=0pt\relax}
\providecommand{\BIBentryALTinterwordstretchfactor}{4}
\providecommand{\BIBentryALTinterwordspacing}{\spaceskip=\fontdimen2\font plus
\BIBentryALTinterwordstretchfactor\fontdimen3\font minus \fontdimen4\font\relax}
\providecommand{\BIBforeignlanguage}[2]{{%
\expandafter\ifx\csname l@#1\endcsname\relax
\typeout{** WARNING: IEEEtran.bst: No hyphenation pattern has been}%
\typeout{** loaded for the language `#1'. Using the pattern for}%
\typeout{** the default language instead.}%
\else
\language=\csname l@#1\endcsname
\fi
#2}}
\providecommand{\BIBdecl}{\relax}
\BIBdecl

\bibitem{pereira2013risk}
A.~A. Pereira, J.~Binney, G.~A. Hollinger, and G.~S. Sukhatme, ``Risk-aware path planning for autonomous underwater vehicles using predictive ocean models,'' \emph{Journal of Field Robotics}, vol.~30, no.~5, pp. 741--762, 2013.

\bibitem{Cowan1988AutomaticSensor}
C.~Cowan and P.~Kovesi, ``{Automatic sensor placement from vision task requirements},'' \emph{IEEE Transactions on Pattern Analysis and Machine Intelligence}, vol.~10, no.~3, pp. 407--416, 1988.

\bibitem{bajcsy2018revisiting}
R.~Bajcsy, Y.~Aloimonos, and J.~K. Tsotsos, ``{Revisiting active perception},'' \emph{Autonomous Robots}, vol.~42, pp. 177--196, 2018.

\bibitem{Placed2023ActiveSLAMSurvey}
J.~A. Placed, J.~Strader, H.~Carrillo, N.~Atanasov, V.~Indelman, L.~Carlone, and J.~A. Castellanos, ``{A Survey on Active Simultaneous Localization and Mapping: State of the Art and New Frontiers},'' \emph{IEEE Transactions on Robotics}, vol.~39, no.~3, pp. 1686--1705, 2023.

\bibitem{Burgard1997ActiveLocalization}
W.~Burgard, D.~Fox, and S.~Thrun, ``{Active mobile robot localization},'' in \emph{Proceedings of the Fifteenth International Joint Conference on Artifical Intelligence - Volume 2}, ser. IJCAI'97.\hskip 1em plus 0.5em minus 0.4em\relax San Francisco, CA, USA: Morgan Kaufmann Publishers Inc., 1997, p. 1346–1352.

\bibitem{Borghi1998MinimumUncertainty}
G.~Borghi and V.~Caglioti, ``{Minimum uncertainty explorations in the self-localization of mobile robots},'' \emph{IEEE Transactions on Robotics and Automation}, vol.~14, no.~6, pp. 902--911, 1998.

\bibitem{Mostegel2014ActiveMonocular}
C.~Mostegel, A.~Wendel, and H.~Bischof, ``{Active monocular localization: Towards autonomous monocular exploration for multirotor MAVs},'' in \emph{2014 IEEE International Conference on Robotics and Automation (ICRA)}, 2014, pp. 3848--3855.

\bibitem{Otsu2018Where}
K.~Otsu, A.-A. Agha-Mohammadi, and M.~Paton, ``{Where to Look? Predictive Perception With Applications to Planetary Exploration},'' \emph{IEEE Robotics and Automation Letters}, vol.~3, no.~2, pp. 635--642, 2018.

\bibitem{Gottipati2019DAL}
S.~K. Gottipati, K.~Seo, D.~Bhatt, V.~Mai, K.~Murthy, and L.~Paull, ``{Deep Active Localization},'' \emph{IEEE Robotics and Automation Letters}, vol.~4, no.~4, pp. 4394--4401, 2019.

\bibitem{Strader2020PerceptionAware}
\BIBentryALTinterwordspacing
J.~Strader, K.~Otsu, and A.-a. Agha-mohammadi, ``{Perception-aware autonomous mast motion planning for planetary exploration rovers},'' \emph{Journal of Field Robotics}, vol.~37, no.~5, pp. 812--829, 2020. [Online]. Available: \url{https://onlinelibrary.wiley.com/doi/abs/10.1002/rob.21925}
\BIBentrySTDinterwordspacing

\bibitem{Tallamraju2020AirCapRL}
R.~Tallamraju, N.~Saini, E.~Bonetto, M.~Pabst, Y.~T. Liu, M.~J. Black, and A.~Ahmad, ``{AirCapRL: Autonomous Aerial Human Motion Capture Using Deep Reinforcement Learning},'' \emph{IEEE Robotics and Automation Letters}, vol.~5, no.~4, pp. 6678--6685, 2020.

\bibitem{Williams2023WhereAmI}
T.~Williams, P.-L. Chen, S.~Bhogavilli, V.~Sanjay, and P.~Tokekar, ``{Where Am I Now? Dynamically Finding Optimal Sensor States to Minimize Localization Uncertainty for a Perception-Denied Rover},'' in \emph{2023 International Symposium on Multi-Robot and Multi-Agent Systems (MRS)}, 2023, pp. 207--213.

\bibitem{Sasaki2020WhereToMap}
T.~Sasaki, K.~Otsu, R.~Thakker, S.~Haesaert, and A.-a. Agha-mohammadi, ``{Where to Map? Iterative Rover-Copter Path Planning for Mars Exploration},'' \emph{IEEE Robotics and Automation Letters}, vol.~5, no.~2, pp. 2123--2130, 2020.

\bibitem{Dhami2023PredNBV}
H.~Dhami, V.~D. Sharma, and P.~Tokekar, ``{Pred-NBV: Prediction-Guided Next-Best-View Planning for 3D Object Reconstruction},'' in \emph{2023 IEEE/RSJ International Conference on Intelligent Robots and Systems (IROS)}, 2023, pp. 7149--7154.

\bibitem{Mac2016HeuristicApproaches}
\BIBentryALTinterwordspacing
T.~T. Mac, C.~Copot, D.~T. Tran, and R.~{De Keyser}, ``{Heuristic approaches in robot path planning: A survey},'' \emph{Robotics and Autonomous Systems}, vol.~86, pp. 13--28, 2016. [Online]. Available: \url{https://www.sciencedirect.com/science/article/pii/S0921889015300671}
\BIBentrySTDinterwordspacing

\bibitem{Karur2021SurveyPathPlanning}
\BIBentryALTinterwordspacing
K.~Karur, N.~Sharma, C.~Dharmatti, and J.~E. Siegel, ``{A Survey of Path Planning Algorithms for Mobile Robots},'' \emph{Vehicles}, vol.~3, no.~3, pp. 448--468, 2021. [Online]. Available: \url{https://www.mdpi.com/2624-8921/3/3/27}
\BIBentrySTDinterwordspacing

\bibitem{Tan2010Navigation}
F.~Tan, J.~Yang, J.~Huang, T.~Jia, W.~Chen, and J.~Wang, ``{A navigation system for family indoor monitor mobile robot},'' in \emph{2010 IEEE/RSJ International Conference on Intelligent Robots and Systems}, 2010, pp. 5978--5983.

\bibitem{Trulls2011AutonomousNavigation}
\BIBentryALTinterwordspacing
E.~Trulls, A.~Corominas~Murtra, J.~Pérez-Ibarz, G.~Ferrer, D.~Vasquez, J.~M. Mirats-Tur, and A.~Sanfeliu, ``{Autonomous navigation for mobile service robots in urban pedestrian environments},'' \emph{Journal of Field Robotics}, vol.~28, no.~3, pp. 329--354, 2011. [Online]. Available: \url{https://onlinelibrary.wiley.com/doi/abs/10.1002/rob.20386}
\BIBentrySTDinterwordspacing

\bibitem{Claes2012Collision}
D.~Claes, D.~Hennes, K.~Tuyls, and W.~Meeussen, ``{Collision avoidance under bounded localization uncertainty},'' in \emph{2012 IEEE/RSJ International Conference on Intelligent Robots and Systems}, 2012, pp. 1192--1198.

\bibitem{Hennes2012Multirobot}
D.~Hennes, D.~Claes, W.~Meeussen, and K.~Tuyls, ``{Multi-robot collision avoidance with localization uncertainty},'' in \emph{Proceedings of the 11th International Conference on Autonomous Agents and Multiagent Systems - Volume 1}, ser. AAMAS '12.\hskip 1em plus 0.5em minus 0.4em\relax Richland, SC: International Foundation for Autonomous Agents and Multiagent Systems, 2012, p. 147–154.

\bibitem{Corominas2020AutonomousNavigation}
A.~Corominas~Murtra, E.~Trulls, O.~Sandoval, J.~Pérez-Ibarz, D.~Vasquez, J.~M. Mirats-Tur, M.~Ferrer, and A.~Sanfeliu, ``{Autonomous navigation for urban service mobile robots},'' in \emph{2010 IEEE/RSJ International Conference on Intelligent Robots and Systems}, 2010, pp. 4141--4146.

\bibitem{Aghamohammadi2014FIRMSF}
A.~akbar Agha-mohammadi, S.~Chakravorty, and N.~M. Amato, ``{FIRM: Sampling-based feedback motion-planning under motion uncertainty and imperfect measurements},'' \emph{The International Journal of Robotics Research}, vol.~33, pp. 268 -- 304, 2014.

\bibitem{Agha2014Robust}
A.-a. Agha-mohammadi, S.~Agarwal, A.~Mahadevan, S.~Chakravorty, D.~Tomkins, J.~Denny, and N.~M. Amato, ``{Robust online belief space planning in changing environments: Application to physical mobile robots},'' in \emph{2014 IEEE International Conference on Robotics and Automation (ICRA)}, 2014, pp. 149--156.

\bibitem{Aghamohammadi2018SLAPSL}
\BIBentryALTinterwordspacing
A.~akbar Agha-mohammadi, S.~Agarwal, S.-K. Kim, S.~Chakravorty, and N.~M. Amato, ``{SLAP: Simultaneous Localization and Planning Under Uncertainty via Dynamic Replanning in Belief Space},'' \emph{IEEE Transactions on Robotics}, vol.~34, pp. 1195--1214, 2018. [Online]. Available: \url{https://api.semanticscholar.org/CorpusID:52934198}
\BIBentrySTDinterwordspacing

\bibitem{vandenBerg2017MotionPlanning}
\BIBentryALTinterwordspacing
J.~van~den Berg, S.~Patil, and R.~Alterovitz, \emph{{Motion Planning Under Uncertainty Using Differential Dynamic Programming in Belief Space}}.\hskip 1em plus 0.5em minus 0.4em\relax Cham: Springer International Publishing, 2017, pp. 473--490. [Online]. Available: \url{https://doi.org/10.1007/978-3-319-29363-9_27}
\BIBentrySTDinterwordspacing

\bibitem{Lee2018MonteCarloTS}
\BIBentryALTinterwordspacing
J.~Lee, G.-h. Kim, P.~Poupart, and K.-E. Kim, ``{Monte-Carlo Tree Search for Constrained POMDPs},'' in \emph{Advances in Neural Information Processing Systems}, S.~Bengio, H.~Wallach, H.~Larochelle, K.~Grauman, N.~Cesa-Bianchi, and R.~Garnett, Eds., vol.~31.\hskip 1em plus 0.5em minus 0.4em\relax Curran Associates, Inc., 2018. [Online]. Available: \url{https://proceedings.neurips.cc/paper_files/paper/2018/file/54c3d58c5efcf59ddeb7486b7061ea5a-Paper.pdf}
\BIBentrySTDinterwordspacing

\bibitem{Lauri2023Partially}
\BIBentryALTinterwordspacing
M.~Lauri, D.~Hsu, and J.~Pajarinen, ``{Partially Observable Markov Decision Processes in Robotics: A Survey},'' \emph{IEEE Transactions on Robotics}, vol.~39, no.~1, p. 21–40, Feb. 2023. [Online]. Available: \url{http://dx.doi.org/10.1109/TRO.2022.3200138}
\BIBentrySTDinterwordspacing

\bibitem{Sun2015HighFrequency}
W.~Sun, S.~Patil, and R.~Alterovitz, ``{High-Frequency Replanning Under Uncertainty Using Parallel Sampling-Based Motion Planning},'' \emph{IEEE Transactions on Robotics}, vol.~31, no.~1, pp. 104--116, 2015.

\bibitem{zheng2020pomdp_py}
\BIBentryALTinterwordspacing
K.~Zheng and S.~Tellex, ``{pomdp\_py: A Framework to Build and Solve POMDP Problems},'' in \emph{ICAPS 2020 Workshop on Planning and Robotics (PlanRob)}, 2020, arxiv link: "\url{https://arxiv.org/pdf/2004.10099.pdf}". [Online]. Available: \url{https://icaps20subpages.icaps-conference.org/wp-content/uploads/2020/10/14-PlanRob_2020_paper_3.pdf}
\BIBentrySTDinterwordspacing

\bibitem{Tange2018}
O.~Tange, \emph{GNU Parallel 2018}.\hskip 1em plus 0.5em minus 0.4em\relax Ole Tange, Mar 2018.

\bibitem{williams2019learning}
T.~Williams and Y.~Sun, ``{Learning State-Dependent, Sensor Measurement Models for Localization},'' in \emph{2019 IEEE/RSJ International Conference on Intelligent Robots and Systems (IROS)}, 2019, pp. 3090--3097.

\bibitem{williams2021learningA}
------, ``{Learning State-Dependent Sensor Measurement Models with Limited Sensor Measurements},'' in \emph{2021 IEEE/RSJ International Conference on Intelligent Robots and Systems (IROS)}, 2021, pp. 86--93.

\bibitem{shek2024localizeriskconstrainedreinforcementlearning}
\BIBentryALTinterwordspacing
C.~L. Shek, K.~Torshizi, T.~Williams, and P.~Tokekar, ``{When to Localize? A Risk-Constrained Reinforcement Learning Approach},'' 2024. [Online]. Available: \url{https://arxiv.org/abs/2411.02788}
\BIBentrySTDinterwordspacing

\end{thebibliography}

\end{document}